\documentclass[runningheads]{llncs}

\usepackage{eccv}

\usepackage{eccvabbrv}

\usepackage{algorithm}
\usepackage{algorithmic}
\usepackage{colortbl}
\usepackage{xcolor}
\usepackage{array}
\usepackage{booktabs}       
\usepackage{tabularx}
\usepackage{amsfonts}       
\usepackage{multirow}
\usepackage{enumitem}
\usepackage{microtype}
\usepackage{graphicx}

\usepackage{url}
\usepackage{bbm}
\usepackage{amsmath, amssymb, mathtools}
\usepackage{wrapfig}
\usepackage{multicol}
\usepackage{bbding}

\usepackage{amsthm}

\usepackage{graphicx}
\usepackage{booktabs}
\usepackage{bbding}
\usepackage{makecell}
\usepackage{stfloats}

\newcommand{\Lapl}{\mathbf{\mathop{\mathcal{L}}}}

\newcommand{\Space}[1]{\mathbb{#1}}
\newcommand{\Set}[1]{\mathcal{#1}}

\newcommand{\wh}[1]{{\color{black}{#1}}}
\newcommand{\hb}[1]{{\color{black}{#1}}}

\usepackage[accsupp]{axessibility}  

\usepackage{authblk}
\usepackage{footmisc}
\usepackage{orcidlink}
\usepackage{hyperref}

\begin{document}

\title{Disentangling Masked Autoencoders for Unsupervised Domain Generalization} 

\titlerunning{DisMAE for UDG}

\author{An Zhang 
\inst{1} \thanks{These authors contribute equally to this work.} \thanks{An Zhang is the corresponding author.}  \orcidlink{0000-0003-1367-711X} 
\and
Han Wang\inst{2}$^*$ \orcidlink{0009-0001-1108-8349} \and
Xiang Wang \inst{3}\orcidlink{0000-0002-6148-6329} \and
Tat-Seng Chua\inst{1}\orcidlink{0000-0001-6097-7807}}

\authorrunning{A.~Zhang et al.}


\institute{National University of Singapore \and Zhejiang University  \and
University of Science and Technology of China \\
\email{anzhang@u.nus.edu}, \email{hwang1024@126.com} \\
\email{xiangwang1223@gmail.com}, \email{dcscts@nus.edu.sg}
}

\maketitle

\begin{abstract}
  Domain Generalization (DG), designed to enhance out-of-distribution (OOD) generalization, is all about learning invariance against domain shifts utilizing sufficient supervision signals.
Yet, the scarcity of such labeled data has led to the rise of unsupervised domain generalization (UDG) — a more important yet challenging task in that models are trained across diverse domains in an unsupervised manner and eventually tested on unseen domains. 
UDG is fast gaining attention but is still far from well-studied.

To close the research gap, we propose a novel learning framework designed for UDG, termed the \underline{Dis}entangled \underline{M}asked \underline{A}uto\underline{E}ncoder (\textbf{DisMAE}), aiming to discover the disentangled representations that faithfully reveal the intrinsic features and superficial variations without access to the class label. 
At its core is the distillation of domain-invariant semantic features, which can not be distinguished by domain classifier, while filtering out the domain-specific variations (for example, color schemes and texture patterns) that are unstable and redundant.   
Notably, DisMAE co-trains the asymmetric dual-branch architecture with semantic and lightweight variation encoders, offering dynamic data manipulation and representation level augmentation capabilities. 
Extensive experiments on four benchmark datasets (\ie DomainNet, PACS, VLCS, Colored MNIST) with both DG and UDG tasks demonstrate that DisMAE can achieve competitive OOD performance compared with the state-of-the-art DG and UDG baselines, which shed light on potential research line in improving the generalization ability with large-scale unlabeled data. Our codes are available at \url{https://github.com/rookiehb/DisMAE}.


\end{abstract}
\section{Introduction}
Domain Generalization (DG) strives to achieve robustness against domain shifts by leveraging high-quality supervision signals \cite{OoD-Bench, DG_survey, ood_theory}. 
However, data is never sufficient for today's large foundation models, and the demands for large-scale labeled source data considerably hinder the breakthrough toward truly generalized models \cite{EqInv}.
Thus, we delve into a more practical yet challenging task - Unsupervised Domain Generalization (UDG) \cite{DARLING}.
Distinct from DG, UDG focuses on learning representations that generalize well from source domains to unseen domains without relying on class labels.
\begin{figure}[t]
    \centering
    \includegraphics[width=0.98\linewidth]{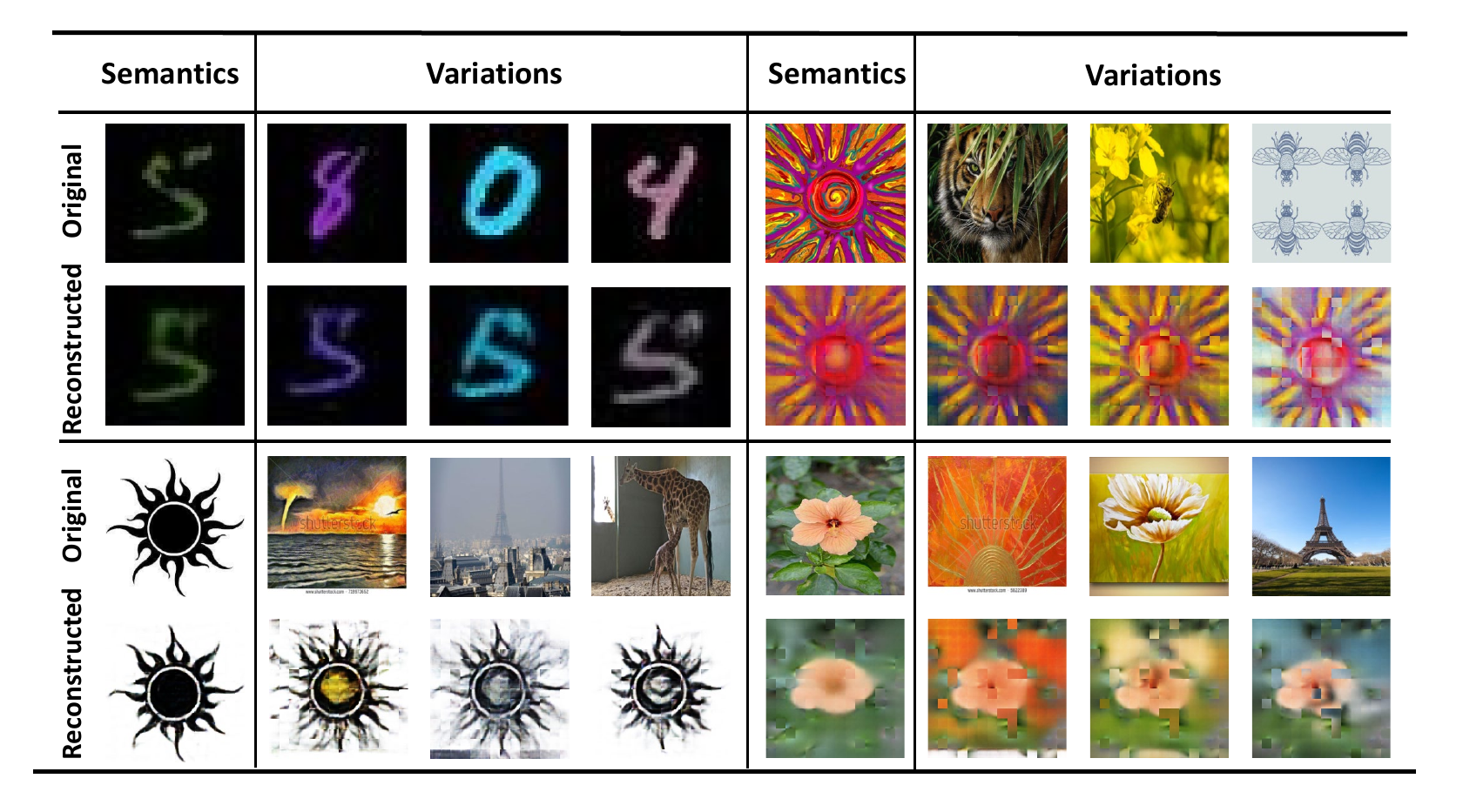}
    \vspace{-10pt}
    \caption{Illustrative reconstructed images generated by DisMAE.
    Rows 1 and 3 present inputs retaining either semantic or variation attributes, sourced from the ColorMNIST and three distinct domains of the DomainNet for a comprehensive comparison.
    Rows 2 and 4 display images crafted by augmenting either original or alternate image variation representations in feature space.
    The evident integration of colors, textures, and backgrounds in these reconstructed images highlights the disentangling capability of DisMAE.
    More examples can be found in Appendix \ref{sec:app_recon}.
    }
    \label{fig:intro}
    \vspace{-10pt}
\end{figure}

We argue that the transition from DG to UDG is non-trivial.
Cutting-edge DG methods, which have shown great success in supervised data, are often ineffective in the UDG task \cite{DomainBed}. 
These methods largely depend on inductive biases derived from available labels \cite{inductive_bias}, leading to suboptimal performance compared to self-supervised learning frameworks such as MoCo \cite{MoCo_V2}, MAE \cite{MAE}, and SimCLR \cite{SimCLR} without supervision. 
Notably, models trained under UDG using unlabeled heterogeneous data can be easily repurposed for DG tasks by assembling classification loss (as elaborated in the methodology section).
Given these observations, our study emphasizes UDG's potential, a research direction that remains largely under-explored.

In light of these challenges and opportunities of UDG, we aim to distill domain-invariant semantic features that faithfully reflect intrinsic properties of the data, ensuring generalization to unseen distributions in an unsupervised manner.
A notable challenge is the entanglement of these features with domain-specific variations such as color schemes and texture patterns \cite{feature_decoupling}. 
Recent advances in DG research hint at a promising direction: the decoupling of domain-invariant attributes from these peripheral variation factors \cite{mDSDI, DSR, VDN}. 
However, with no access to the class label, ensuring domain-invariant features truly encapsulate inherent data attributes poses great challenges \cite{DDG, DecAug}. 
Furthermore, in some real-world scenarios, distribution shifts among source domains overshadow intra-domain class differences \cite{NICO}.
For instance, the transformation from a photo to a sketch of a Teddy manifests more significant changes than transitioning to a Bichon.
This strong heterogeneity often results in representations emphasizing broad distributional shifts over nuanced domain-specific variations.

Toward this end, we propose a new learning framework designed for UDG, \underline{Dis}entangled \underline{M}asked \underline{A}uto\underline{E}ncoder (\textbf{DisMAE}), that integrates invariance and disentanglement principles.
By the disentanglement principle, we mean that the representations are decomposed into two components - semantic features and variations.
To force the semantic representations to capture less redundant information and thus preferably disentangle the variations, we pull the reconstructed samples with their original semantics and variations closer to the input and push the reconstructed samples with different in-domain variations apart.
By invariance principle, we mean that the semantic representations are invariant throughout a variety of domain changes, therefore their domain category is undistinguished.
Put simply, variations should encapsulate domain-specific features, leaving the disentangled semantic representations to exclusively preserve intrinsic information.

Guided by these two principles, our DisMAE strategy incorporates four modules: a transformer-based semantic encoder, a lightweight variation encoder, a transformer-based decoder, and a domain label-enhanced invariance classifier. 
For the reconstruction part, the asymmetric dual-branch architecture with two separate encoders and one single decoder is implemented together to avoid information collapse. 
We additionally introduce an adaptive contrastive loss to keep two branches capturing information in a principled disentangled manner.
Specifically, towards the disentanglement principle, target semantic representations are concatenated with other intra-domain samples' variations, ensuring the reconstructed samples diverge from the original input.
Towards the invariance principle, the domain label-enhanced invariance classifier predicts the probability of correctly identifying the domain category, and these probabilities are then inversely applied as the adaptive weights to re-weight the contrastive loss.
Jointly training under these two principles enables the DisMAE models to disentangle the domain-invariant semantic features and domain-specific variations, and further boosts the model's capability in enriching the variation through latent space augmentation. 



\section{Preliminary}

The goal of unsupervised domain generalization (UDG) is to learn a domain-invariant feature extractor from diverse domains in a self-supervised manner, thereby enabling robust generalization to the target unseen domain \cite{DiMAE}.
Let $\Set{D}_{\text{train}} = \{x \in \Set{I}_d| d \in\{1,\cdots,M-1\}\}$ be the training set that involves $M-1$ source domains.
Each training domain $\mathcal{I}_{d}$ comprises a collection of unlabeled images $\{x_i\}$.
Let $\Set{D}_{\text{test}} = \{x \in \Set{I}_M\}$ represent the test set containing previously unseen images from the target domain $\mathcal{I}_M$. 

In order to learn a domain-invariant feature extractor $\phi_s$ in self-supervised learning, there is a foundation assumption \cite{IP-IRM}: any given sample $x$ is generated from two disentangled features, $x=g(\mathbf{s}_x,\mathbf{v}_x)$, where $\mathbf{s}_x$ denotes the semantic representation that invariant to domain shifts, $\mathbf{v}_x$ signifies the superficial variations that change across domains.
The generative function $g(\cdot,\cdot)$ maps these two latent space representations back into the sample space.
In particular, this assumption naturally encodes two principles: disentanglement and invariance principles.

\vspace{5pt}
\noindent\textbf{Disentanglement Principle.} \textit{There exists a direct product decomposition of attributes as well as feature representations in real-world visual generation scenarios.}

\begin{wrapfigure}{r}{0.5\textwidth}
\vspace{-23pt}
    \centering
    \includegraphics[width=0.95\linewidth]{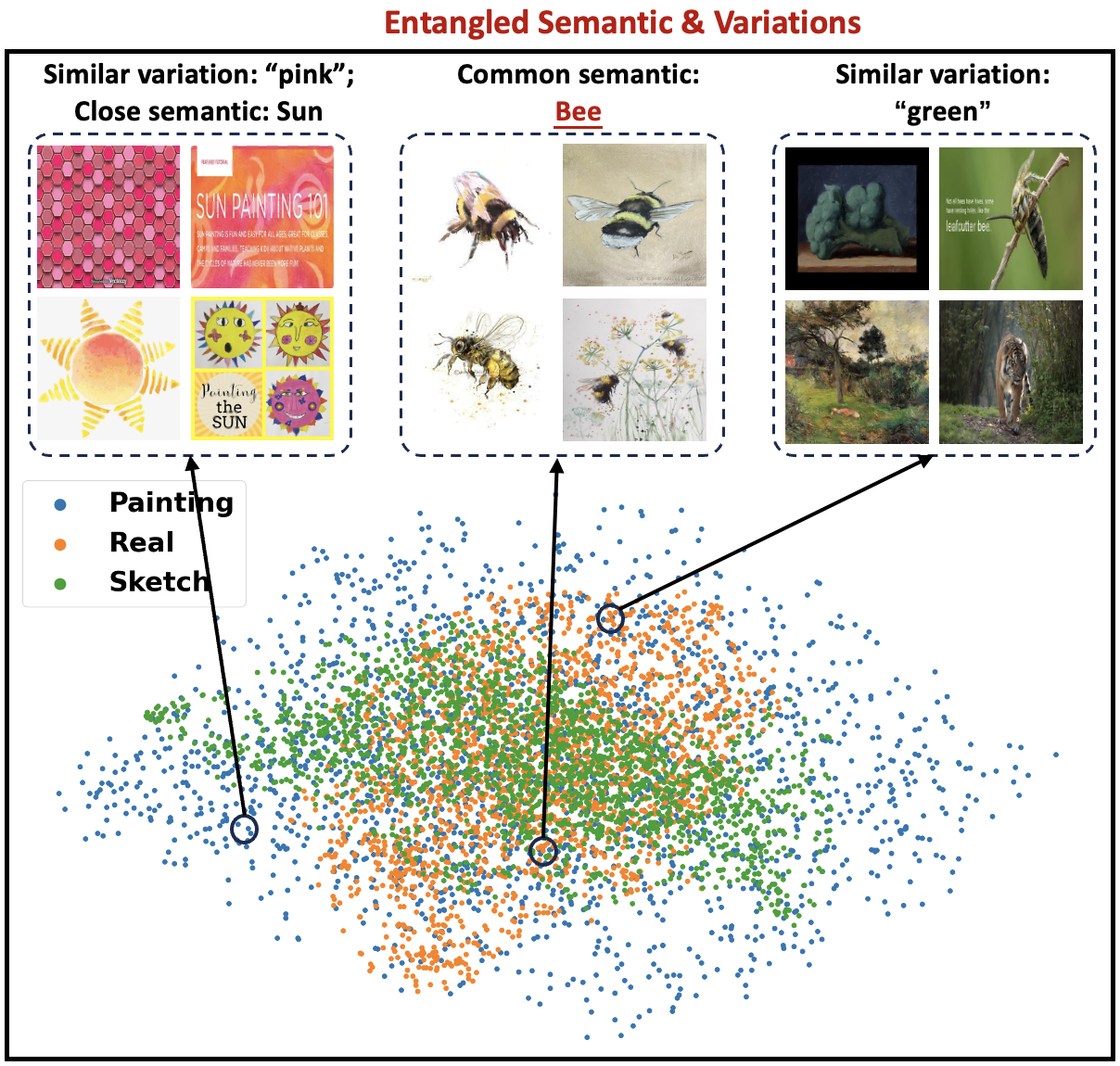}
    \caption{t-SNE visualization of MAE representations on DomainNet. 
    Points are color-coded based on their domain labels. 
    MAE demonstrates inconsistent feature spaces across different domains, stemming from capturing entangled features of both semantics and variations.
    }
    \label{fig:tsne_MAE}
    \vspace{-10pt}
\end{wrapfigure}

Let $s_x$ be the semantic attribute of sample $x$, and $v_x$ denote its variations, \eg for an image in the Colored MNIST dataset \cite{IRM}, its semantic attribute $s_x$ signifies ``digit'', while the variation attributes $v_x$ is ``color''.
There exists a direct product decomposition of attributes that could control each sample $x$'s generation: $s_x \times v_x$, \eg generating a digit ``5'' with ``green'' color (see examples in Figure \ref{fig:intro}).
Thus, this decomposition implies a disentangled representation learning, $\phi_s \times \phi_v$, that maps image pixels to a representation space.
Here, $\mathbf{s}_x:= \phi_s(x)$ captures the semantic aspects, and $\mathbf{v}_x:=\phi_v(x)$ encapsulates the variations of sample $x$.

The disentanglement principle is inherent in a desirable property. 
Modifying attributes is essentially akin to directly altering representations in the feature space. 
For instance, consider the scenario illustrated in Figure \ref{fig:intro}, where transforming a digit ``5'' from being ``green'' to ``purple'' in the attributes space corresponds to the alteration of ``5'' $\times$ ``green'' to ``5'' $\times$ ``purple''.
This change can be readily accomplished by substituting the variation representation $\phi_v(x)$.
Thus, through the lens of the disentanglement principle, we highlight the potential to diversify variations using representation-level data augmentation.

\vspace{5pt}
\noindent\textbf{Invariance Principle.} 
\textit{The ideal UDG model refines domain-invariant representations that causally determine the intrinsic attributes, regardless of changes in domains.}
\vspace{1pt}

In other words, the domain-invariant feature extractor $\phi_s$ should exclude any domain-specific features identifiable by a domain classifier. 
More formally, let us consider $x_i \in \Set{I}_{d_1}$ and $x_j \in \Set{I}_{d_2}$. For these cases, the support of semantic representations for domain $d_1$ should align entirely with that for domain $d_2$, implying $\text{Sup}(\phi_s(x_i)) = \text{Sup}(\phi_s(x_j))$, for all $i,j$.
Here $\text{Sup}(\cdot)$ denotes the support of distribution.

Failing to adhere to the invariance principle can lead to entangled representation learning, which ultimately undermines the ability for effective generalization. 
As depicted in Figure \ref{fig:tsne_MAE}, the widely used self-supervised learner MAE tends to learn representation spaces that incompletely overlap across domains. 
This observation underscores the idea that self-supervised learners tend to faithfully capture distinctions among various domain data by indiscriminately extracting both semantic and variation features.

\vspace{5pt}
\noindent\textbf{Discussion.}
These two principles play a crucial role in acquiring a robust semantic extractor $\phi_s$ in UDG.
Through the disentangled representations learning, even when testing a new sample from an unseen domain, the domain-invariant semantics of the sample have already been learned as features during training. 
As a result, the feature extractor trained on known domains remains applicable. 
Moreover, the new combination of semantics and variations can reshape the existing sample distribution, consequently boosting the model's generalization capability through representation-level data augmentation.
\section{Methodology}
\begin{figure*}[t]
    \centering
    \includegraphics[width=0.98\linewidth]{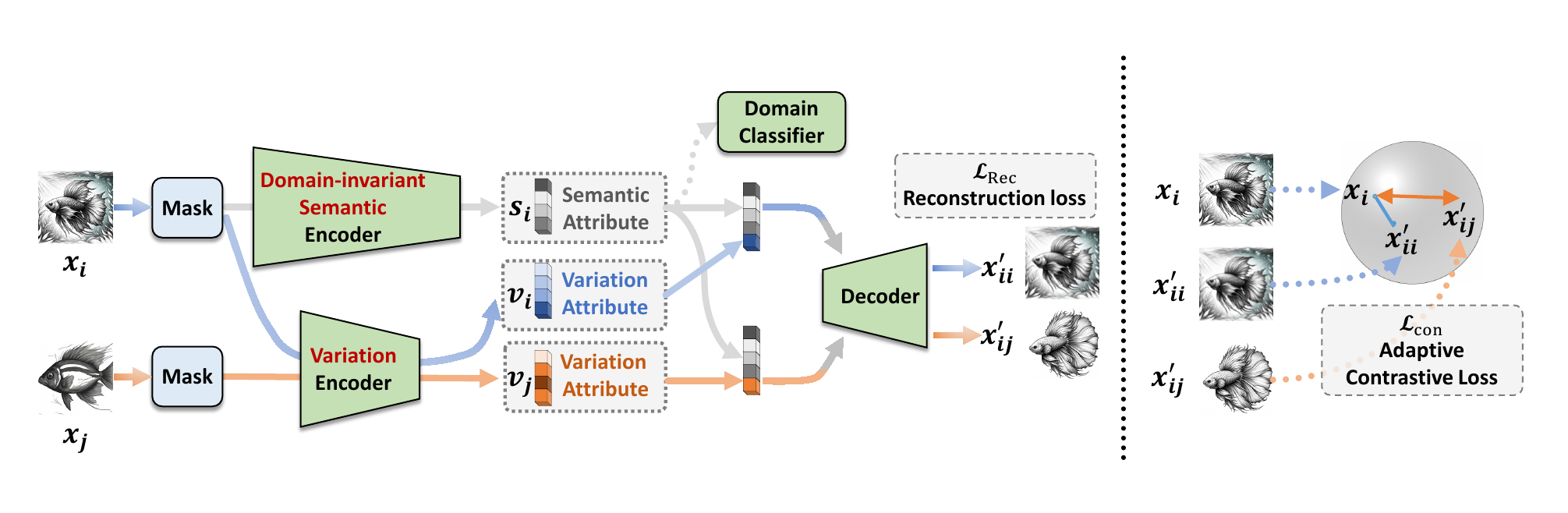}
    \vspace{-10pt}
    \caption{Framework of our proposed DisMAE. DisMAE develops an asymmetric dual-branch architecture, with the upper main branch distilling the domain-invariant semantics, along with the lightweight branch extracting the domain-specific variations.
    A domain classifier is integrated to quantify the degree of domain-specific information embedded within the semantic encoder, thereby monitoring its acquisition of domain-invariant knowledge. Note that the domain classifier is updated while freezing backbones and is only used for generating adaptive weights.}
    \label{fig:framework}
    \vspace{-10pt}
\end{figure*}
We now introduce our Disentangled Masked Autoencoder (DisMAE), designed to derive decomposed high-level representations for UDG, building on the scalable self-supervised MAE framework \cite{MAE}. 
Grounded in disentanglement and invariance principles, our approach integrates reconstruction loss and adaptive contrastive loss in a collaborative manner to guide the learning process.
An overview of DisMAE's framework is provided in Figure \ref{fig:framework}.

\subsection{DisMAE}
Considering a multi-domain reconstruction task with the training set $\Set{D}_{\text{train}} = \{ (x_i, d_i)\}_{i=1}^N$, where $x_i \in \Set{I}_{d_i}$ denotes the input image, $d_i$ signifies the corresponding domain category, and $N$ represents the number of training data.
DisMAE develops an asymmetric dual-branch architecture, visualized in Figure \ref{fig:framework}, with a main branch mapping the input image into the semantic feature space \ie semantic encoder $\phi_s(\cdot)$, along with a lightweight branch that extracts the input sample's variation attributes \ie variations encoder $\phi_v(\cdot)$.
In alignment with MAE \cite{MAE}, \wh{we denote the masking mechanism as the function $\mathcal{M}$ and the input visible patches as $\mathcal{M}(x_i)$ for the original image $x_i$}. Our approach utilizes two transformer-based encoders \cite{ViT} to intensify the domain-invariant semantics and domain-specific variations as representations:
\begin{equation}
    (\mathbf{s}_{i}, \mathbf{v}_{i}) = (\phi_s(\mathcal{M}(x_i)),\phi_v(\mathcal{M}(x_i))).
\end{equation}
Consequently, the domain-invariant semantic representation $\mathbf{s}_i$ and domain-specific variation representation $\mathbf{v}_i$ are to estimate the attributes $s_{x_i}$ and $v_{x_i}$ in feature space, respectively. \wh{We denote the [cls] patch embedding of $\mathbf{s}_i$ and $\mathbf{v}_i$ as $\mathbf{s}_i^0$ and $\mathbf{v}_i^0$, where $\mathbf{s}_i^0, \mathbf{v}_i^0 \in \mathbb{R}^{B \times H}$ and $B$ is the batch size and $H$ is the hidden size.}
Their combination can parameterize image $x_i$ as:
\begin{equation}
    \mathbf{x}_{ii} = \mathbf{s}_i || \mathbf{v}_i^0,
\end{equation}
where $||$ symbolizes the concatenation operation. \wh{Here, we utilize $\mathbf{v}_i^0$ since we only need overall information about variation features.}

In addition to the encoders, we utilize a transformer-based decoder, denoted as $g(\cdot, \cdot)$, mirroring the architecture of the MAE decoder.
This enables the reconstruction $x_{ii}' = g(\mathbf{s}_i, \mathbf{v}_i^0)$ of $x_i$.
Importantly, the decoder is operational only during training for reconstruction. 
During the UDG task's fine-tuning phase, solely the main branch, namely the semantic encoder $\phi_s$, is deployed.

For the effective optimization of both encoders and the decoder, the prevalent reconstruction learning strategy is employed.
Specifically, the risk function measures the quality of reconstructed pixel values, which can be formulated as the mean squared error (MSE) or $\gamma$-constrained reconstruction loss \cite{DDG} \wh{between the masked patches and the corresponding masked patches for the original image $x_i$}.
Here we apply $\gamma$-constrained reconstruction loss, which is defined as:
\begin{equation}\label{eq:loss_rec}
    \Lapl_{\text{rec}} = \frac{1}{N}\sum_{i=1}^{N}\max \{||x_i - g(\mathbf{s}_i, \mathbf{v}_i^0)||_{l_2} - \gamma, \, 0\}, 
\end{equation}
where $\gamma > 0$ serves as a hyperparameter, indicating the bound of the allowable reconstruction error. 
However, solely minimizing the risks over the original data distribution fails to model domain-invariant information and suffers from poor domain generalization.


\subsection{Implementation of Two Principles}
To bring forth better generalizations \wrt domain shift, we advocate for disentanglement and invariance principles.
To parameterize the disentanglement principle, we devise an augmentation operator on semantic and variation representations, which preserves the estimated domain-invariant intrinsic attributes but intervenes in the estimated domain-specific information.
Formally, the operator samples in-batch variation representations $\mathbf{v}_j^0$ from distinct sample $x_j$ to replace $\mathbf{v}_i^0$ of $x_i$ and later combine it with $\mathbf{s}_i$ to generate the augmented image $x_{ij}'$.
The augmentation process is:
\begin{equation}
   \mathbf{x}_{ij} = \mathbf{s}_i || \mathbf{v}_j^0, \quad x_{ij}' = g(\mathbf{s}_i, \mathbf{v}_j^0).
\end{equation}

Having established the augmented representations, we enforce the reconstructed sample with its original semantics and variations closer to the input sample while concurrently pushing the reconstructed samples possessing distinct intra-domain variations away from it.
This can be captured by a contrastive loss between the original image $x_i$ and its reconstruction $x_{ii}'$:
\begin{equation} \label{eq:loss_contrastive}
    l(x_i,x_{ii}') = - \log \frac{\exp(s(x_i,x_{ii}')/\tau)}{\sum_{j\in \Set{I}_{d_i}\cup \{i\}} \exp(s(x_i,x_{ij}')/\tau)},
\end{equation}
where $\tau$ is the temperature hyperparameter in the contrastive loss \cite{InfoNCE}, and $s(\cdot,\cdot)$ is a similarity function quantifying the distance between the reconstructed and original images \wh{on the masked patches}.
For this, we adopt the negative $\gamma$-constrained reconstruction loss $s(x_i,x_{ii}')=-\max \{||x_i - x_{ii}'||_{l_2} - \gamma, \, 0\}$ as the similarity metric.
We emphasize that due to the strong heterogeneity of different domain samples, intra-domain negatives $j\in \Set{I}_{d_i}$ are exclusively selected in equation \eqref{eq:loss_contrastive}.

To further refine our approach, we instantiate the invariance principle, enabling DisMAE to learn domain-invariant representations that remain indistinguishable by the domain classifier $f(\cdot)$.
We start by determining the degree of domain-specific information in the semantic representations $\mathbf{s}_i^0$ of image $x_i$ by employing a two-layer multilayer perception (MLP) as the domain classifier, with the cross-entropy loss serving as the objective function:
\begin{equation}
    p(x_i\in \mathcal{I}_{d_i}|\mathbf{s}_i^0) = \text{Softmax}(d_i, \hat{d_i}=f(\mathbf{s}_i^0)).
\end{equation}

Drawing inspiration from Inverse Propensity Weighting (IPW) in causal inference \cite{IPW, pearl2016causal}, we multiply the inverse of this degree to reweight the loss, leading us to our adaptive contrastive loss formulation:
\begin{equation}
    \Lapl_{\text{con}} = \sum_{i=1}^{N} \frac{1}{p(x_i\in \mathcal{I}_{d_i}|\mathbf{s}_i^0)} \cdot l(x_i,x_{ii}').
\end{equation}
This adaptive contrastive loss compels DisMAE to prioritize the refinement of domain-invariant semantic encoder learning, simultaneously mitigating the impact stemming from biased representation learning.

Overall, we can aggregate all the foregoing risks and attain the final objective of DisMAE in the UDG task:
\begin{equation}
    \Lapl_{\text{UDG}} =\Space{E}_{(x_i, d_i)\in\Set{D}_{\text{train}}} (\Lapl_{\text{rec}} + \lambda_1 \Lapl_{\text{con}}),
\end{equation}
with $\lambda_1$ acting as the hyperparameter to control the strength of disentanglement and invariance principles.
In the inference phase, we only use the domain-invariant semantic representations $\mathbf{s}_x^0$ to make predictions, shielding them from the influence of domain shifts.

In the context of domain generalization tasks, the training set is formulated as $\Set{D}_{\text{train}} = \{ (x_i, y_i, d_i)\}_{i=1}^N$, where $y_i$ refers to the class label for image $x_i$ within domain $d_i$.
To adapt DisMAE for domain generalization tasks, it is a straightforward process of introducing an additional cross-entropy loss that utilizes the supervision signal $y_i$.
The overall objective function for this adaptation is formulated as:
\begin{equation}
    \Lapl_{\text{DG}} =\Space{E}_{(x_i, y_i, d_i)\in\Set{D}_{\text{train}}} (\Lapl_{\text{rec}} + \lambda_1 \Lapl_{\text{con}} + \lambda_2 \text{CE}(y_i, \hat{y_i}(\mathbf{s}_i^0))),
\end{equation}
where $\lambda_1$ and $\lambda_2$ are hyperparameters.







\section{Experiments}

We aim to answer the following research questions:
\begin{itemize}
    \item[$\bullet$] \textbf{RQ1}: How does DisMAE perform compared with prevailing UDG and DG approaches?
    \item[$\bullet$] \textbf{RQ2}: Does DisMAE successfully derive disentangled representations and domain-invariant features in UDG scenarios?
    \item[$\bullet$] \textbf{RQ3}: What are the impacts of the components on our DisMAE?
\end{itemize}

\noindent\textbf{Datasets}. 
For a comprehensive comparison, we evaluate DisMAE across both unsupervised domain generalization (UDG) and domain generalization (DG) tasks. 
For UDG, our evaluation is on DomainNet \cite{DomainNet}, a dataset with 586,575 images spanning 345 object classes across six domains: Real, Painting, Sketch, Clipart, Infograph, and Quickdraw. 
Our DG evaluations center on two benchmark datasets: PACS \cite{PACS} and VLCS \cite{VLCS}. 
PACS contains 9,991 images over 7 classes and 4 domains: Art, Cartoons, Photos, and Sketches. 
VLCS includes 10,729 images across 5 classes from four domains: Caltech101, LabelMe, SUN09, and VOC2007.

\vspace{5pt}
\noindent\textbf{Implementation Details}. 
Contrary to the standard use of ResNet-18 \cite{ResNet}, we employ the more powerful self-supervised learner ViT-B/16 \cite{MAE} as our default backbone (for experiments in section \ref{sec:exp_rq3}, ViT-Tiny/16 serves as the backbone). 
We adopt a learning rate of 1e-4, a weight decay of 0.05, and a batch size of $N_d \times 96$, where $N_d$ denotes the number of domains in the training dataset.
For UDG tasks, in alignment with the all correlated setting from DARLING \cite{DARLING}, our model is pretrained on DomainBed, bypassing the ImageNet dataset. 
Conversely, for DG tasks, we initialize our backbone using ImageNet pre-training, as prescribed by DomainBed \cite{DomainBed}. 
Detailed information on implementation and hyperparameter settings can be found in Appendix \ref{sec:app_implementation}.

\vspace{5pt}
\noindent\textbf{Baselines}. 
For UDG tasks, we benchmark DisMAE against notable contrastive learning approaches (MoCo V2 \cite{MoCo_V2}, BYOL \cite{BYOL}, DARLING \cite{DARLING}) and generative-based methods (MAE \cite{MAE}, CycleMAE \cite{cycleMAE}).
For DG tasks, our comparisons encompass diverse learning strategies: from the vanilla (ERM \cite{ERM}), distributional robust optimization (GroupDRO \cite{GroupDRO}), data augmentation-based (Mixup \cite{Mixup}), to domain-invariant learning (IRM \cite{IRM}, MMD \cite{MMD}), variance optimization (VREx \cite{VREx}, Fishr \cite{Fishr}), and disentangled representation learning (DDG \cite{DDG}). 
Related work and detailed implementation are available in Section \ref{sec:app_related_work} and Appendix \ref{sec:app_experiment_settings}.

\subsection{Overall Performance Comparison (RQ1)}

\subsubsection{Evaluations on UDG.}\,\,

\begin{table*}[t]
\centering   
   \resizebox{0.98\linewidth}{!}{
   \begin{tabular}{c c c|c c c|c c|c c c|c c}
   \toprule
   & & & \multicolumn{5}{c|}{\textbf{Label Fraction 1\% (Linear evaluation)}} & \multicolumn{5}{c}{ \textbf{Label Fraction 5\% (Linear evaluation)}} \\
    \midrule
    \multicolumn{3}{c|}{\textbf{Methods}} & \textbf{Clip} & \textbf{Info} & \textbf{Quick}  & \textbf{Overall} & \textbf{Avg.} & \textbf{Clip} & \textbf{Info} & \textbf{Quick} & \textbf{Overall} & \textbf{Avg.} \\
    \midrule
    \multicolumn{3}{c|}{ERM \cite{ERM}} & 9.38 & \textbf{9.70} & 6.64 & 8.04 & 8.57 & 11.27 & 8.58 & 6.90 & 8.25 & 8.92 \\
    \multicolumn{3}{c|}{MoCo V2 \cite{MoCo_V2}} & 9.76 & 6.53 & 5.24 & 6.51 & 7.18 & 11.44 & 9.20 & 4.95 & 7.42 & 8.53 \\
    \multicolumn{3}{c|}{BYOL \cite{BYOL}} & 17.93 & 8.03 & 5.05 & 8.47 & 10.34 & 24.26 & 8.88 & 4.66 & 9.78 & 12.60 \\
    \multicolumn{3}{c|}{MAE \cite{MAE}} & \underline{19.69} & 8.85 & \textbf{12.49} & \underline{12.96} & \underline{13.68} & \underline{29.12} & 8.88 & \underline{13.71} & \underline{15.52} & \underline{17.24} \\
    \multicolumn{3}{c|}{DARLING \cite{DARLING}} & 13.77 & 6.33 & 7.24 & 8.41 & 9.18 & 19.02 & 8.49 & 9.59 & 11.19 & 12.36 \\
    \multicolumn{3}{c|}{CycleMAE \cite{cycleMAE}} & 11.68 & 6.82 & 8.06 & 8.46 & 8.85 & 22.06 & \textbf{10.46} & 12.07 & 13.66 & 14.86 \\
    \midrule
    \multicolumn{3}{c|}{\textbf{DisMAE (Ours)}} & \textbf{24.55} & \underline{9.18} & \underline{11.75} & \textbf{13.64} & \textbf{15.16} & \textbf{31.70} & \underline{10.14} & \textbf{19.06} & \textbf{19.19} & \textbf{20.30} \\
    \midrule
    & & & \multicolumn{5}{c|}{\textbf{Label Fraction 10\% (Full finetuning)}} & \multicolumn{5}{c}{\textbf{Label Fraction 100\% (Full finetuning)}} \\
    \midrule
    \multicolumn{3}{c|}{\textbf{Methods}} & \textbf{Clip} & \textbf{Info} & \textbf{Quick}  & \textbf{Overall} & \textbf{Avg.} & \textbf{Clip} & \textbf{Info} & \textbf{Quick} & \textbf{Overall} & \textbf{Avg.} \\
    \midrule
    \multicolumn{3}{c|}{ERM \cite{ERM}} &27.57&\underline{13.32}&8.29&13.60&16.39&48.61&18.37&20.06&25.38&29.01 \\
    \multicolumn{3}{c|}{MoCo V2 \cite{MoCo_V2}} &26.51&11.68&10.62&14.12&16.27&57.16&16.44&29.40&31.50&34.33 \\
    \multicolumn{3}{c|}{BYOL \cite{BYOL}} &31.33&11.55&9.51&14.48&17.46&56.62&12.77&26.90&29.08&32.10 \\
    \multicolumn{3}{c|}{MAE \cite{MAE}} &\underline{41.76}&12.91&\underline{18.98}&\underline{21.94}&\underline{24.55}&\underline{63.90}&\underline{19.34}&33.00&\underline{35.54}&\underline{38.75} \\
    \multicolumn{3}{c|}{DARLING \cite{DARLING}} &30.68&11.66&13.93&16.69&18.75&55.20&17.04&\underline{33.30}&33.22&35.18 \\
    \multicolumn{3}{c|}{CycleMAE \cite{cycleMAE}} &33.84&12.06&15.41&18.23&20.43&56.10&19.15&28.60&31.60&34.62 \\
    \midrule
    \multicolumn{3}{c|}{\textbf{DisMAE (Ours)}} &\textbf{49.06}&\textbf{15.82}&\textbf{22.68}&\textbf{26.16}&\textbf{29.19}& \textbf{72.99}&\textbf{23.40}&\textbf{38.20}&\textbf{41.22}&\textbf{44.86} \\
    \bottomrule
    \end{tabular}  }  
    \vspace{5pt}
    \caption{Unsupervised domain generalization results on DomainNet with Painting, Real, and Sketch serving as training domains, while Clipart, Infograph, and Quickart for testing.
    Each model undergoes unsupervised pre-training prior to fine-tuning on labeled data. 
    Overall and Avg. are the overall test data accuracy and the arithmetic mean of individual domain accuracy respectively. 
    Note that these metrics differ, given the varying sizes of the test domains.
    \textbf{Bold}=best, \underline{underline}=second best.}
    \vspace{-10pt}
    \label{tab:UDG1}
\end{table*}

\vspace{5pt}
\noindent\textbf{Setting}.
For a balanced comparison, we strictly follow the all-correlated setting outlined by DARLING \cite{DARLING}.
We select 20 classes from DomainNet's \cite{DomainNet} 345 categories for both training and testing phases, using Painting, Real, and Sketch as source domains, while Clipart, Infograph, and Quickdraw act as target domains, and vice versa.
Specifically, the UDG task is executed in three stages:
1. Unsupervised training on the source domains.
2. Depending on the labeled data proportion from the source domain, we either fit a linear classifier on the frozen semantic encoder or fully fine-tune the network.
In particular, for labeled data proportions under 10\% of the source domain's total training set, we opt for fitting a linear classifier. 
If it exceeds 10\%, the entire network undergoes fine-tuning.
3. Model evaluation on the target domains.


\vspace{5pt}
\noindent\textbf{Results}. 
Tables \ref{tab:UDG1} and \ref{tab:UDG2} showcase a performance comparison for the UDG evaluation using the ViT-B/16 backbone.
The best-performing methods per test are bold, while the second-best methods are underlined. 
We observe that: 

\begin{itemize}[leftmargin=*]
    \item \textbf{DisMAE consistently and significantly outperforms all baseline models in terms of both average and overall accuracy.}
    In Table \ref{tab:UDG1}, DisMAE achieves gains of 1.48\%, 3.04\%, 4.65\%, and 6.11\% for average accuracy across the 1\%, 5\%, 10\%, and 100\% fraction settings, respectively. 
    The robustness of DisMAE is attributed to its proficiency in distilling the invariant semantic attributes, regardless of variations in domain-specific features.
    \item \textbf{Entangled baselines exhibit instability when faced with domain shifts.} 
    Conventional contrastive methods display limited generalization capability, revealing vulnerability in handling unknown target domains. 
    These methods rely on data augmentations to form positive and negative pairs, lacking the explicit ability to bridge domain gaps by effectively pulling positive pairs together and pushing negative pairs apart. 
    In the case of DARLING, negative samples are generated for each queue based on the dissimilarity between diverse domain samples. 
    However, the cross-domain pairs constructed in this way are noisy in some cases, ultimately leading to suboptimal performance. 
    Generative-based methods such as CycleMAE employ cycle reconstruction tasks to construct cross-domain pairs while suffering from an implicit domain-variant encoder, resulting in performance deterioration. 
    In contrast, leveraging the strength of a domain-invariant extractor, our DisMAE significantly enhances performance by a substantial margin.
\end{itemize}


\subsubsection{Evaluations on DG.}\,\,

\begin{table}[t]
\centering  
   \resizebox{0.55\linewidth}{!}{
   \begin{tabular}{c c c|c c c c|c}
   \toprule
    \multicolumn{3}{c|}{\textbf{Methods}}& \textbf{C}& \textbf{L} & \textbf{S} &\textbf{V} & \textbf{Avg.} \\
    \midrule 
    \multicolumn{3}{c|}{ERM \cite{ERM}}&95.47&65.74&76.83&77.07&78.78 \\
    \multicolumn{3}{c|}{MMD \cite{MMD}}&94.61&65.85&73.70&77.26&77.86 \\
    \multicolumn{3}{c|}{GroupDRO\cite{GroupDRO}}&95.49&65.80&76.28&77.37&78.74 \\
     \multicolumn{3}{c|}{IRM \cite{IRM}}&96.26&65.30&73.25&76.01&77.71 \\
     \multicolumn{3}{c|}{Mixup\cite{Mixup}}&95.38&65.74&76.83&77.07&78.76 \\
     \multicolumn{3}{c|}{VREx\cite{VREx}}&96.94&63.67&73.34&76.10&77.51 \\
     \multicolumn{3}{c|}{Fishr\cite{Fishr}}&96.17&65.19&74.77&\underline{78.06}&78.55 \\
     \multicolumn{3}{c|}{DDG\cite{DDG}}&\underline{96.97}&64.21&71.76&76.58&77.38 \\
     \multicolumn{3}{c|}{RIDG\cite{RIDG}}&96.64&\underline{66.16}&\textbf{77.57}&76.68&\underline{79.26} \\
     \midrule
     \multicolumn{3}{c|}{\textbf{DisMAE (Ours)}}&\textbf{97.35}&\textbf{66.96}&\underline{77.10}&\textbf{80.12}&\textbf{80.38} \\
     \bottomrule
    \end{tabular}}
    \vspace{5pt}
\caption{Domain generalization results on VLCS. 
}
\label{tab:vlcs}
\vspace{-10pt}
\end{table}

\vspace{5pt}
\noindent\textbf{Setting}.  
We strictly follow DomainBed \cite{DomainBed} evaluation protocols, utilizing training-domain validation set.
One domain serves as the target (test), while the remaining acts as source (training) domains.
Training is done with 80\% of the source data, and validation uses the remaining 20\%. 
We implement a hyperparameter random search, following the DomainBed \cite{DomainBed} guidelines. 
Optimal hyperparameters are chosen based on validation set performance for each test, and the model is subsequently evaluated on the target domain.
To mitigate the influence of randomness, experiments are conducted thrice with distinct seeds, and we report the average accuracy from these runs.

\vspace{5pt}
\noindent\textbf{Results}. 
As illustrated in Tables \ref{tab:vlcs} and \ref{tab:PACS}, surprisingly, DisMAE demonstrates steady superiority over all state-of-the-art DG baselines concerning the average accuracy on VLCS and PACS. 
Specifically, compared to the state-of-the-art baseline, DisMAE achieves substantial gains of 1.12\% and 0.50\% in average accuracy for VLCS and PACS, respectively. 
In contrast, baselines perform unstablely across various sub-tasks. 
Benefiting from disentangled and domain-invariant representations, DisMAE successfully achieves more robust performance.

\subsection{Discussion about Two Principles (RQ2)}

\begin{figure*}[t]
	\centering
	\subcaptionbox{DisMAE semantic representations  $\mathbf{s}_x^0$\label{fig:tsne_DisMAE_s}}{
		\includegraphics[width=0.45\linewidth]{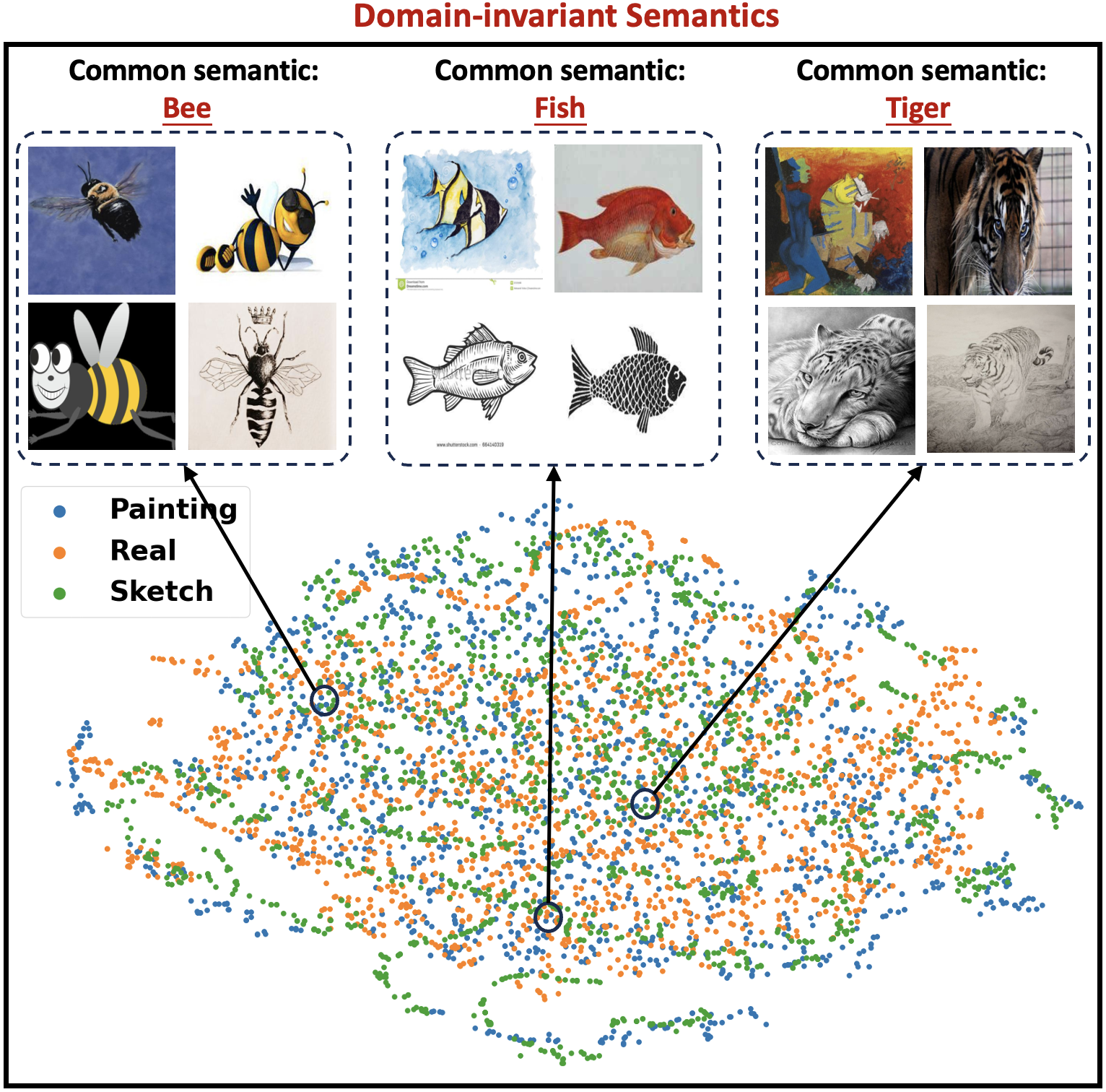}}
    \subcaptionbox{DisMAE variation representations $\mathbf{v}_x^0$\label{fig:tsne_DisMAE_v}}{   
		\includegraphics[width=0.45\linewidth]{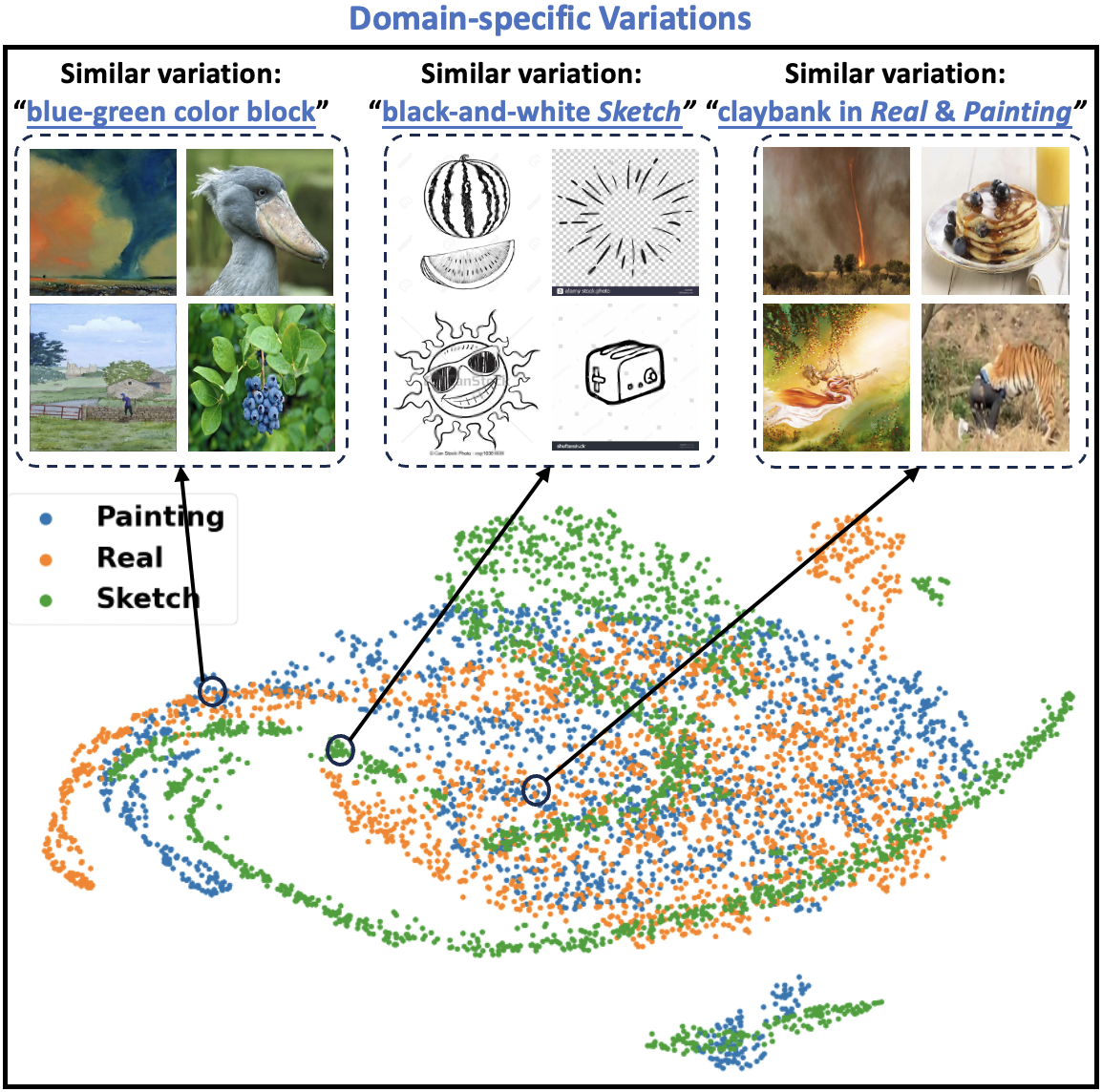}}
	\caption{t-SNE visualization of DisMAE semantic and variation representations on DomainNet, with points color-distinguished by their domain labels. 
    (\ref{fig:tsne_DisMAE_s}) DisMAE semantic representations are mixed and interspersed regardless of domain labels, showcasing the ability of the semantic encoder to discern domain-invariant features.
    (\ref{fig:tsne_DisMAE_v}) Samples from different domains reside on distinct manifolds, highlighting the variation encoder's capability to extract domain-specific features.
    }
	\label{fig:tsne_DisMAE}
    \vspace{-10pt}
\end{figure*}

\begin{figure}[t]
    \centering
    \includegraphics[width=0.65\linewidth]{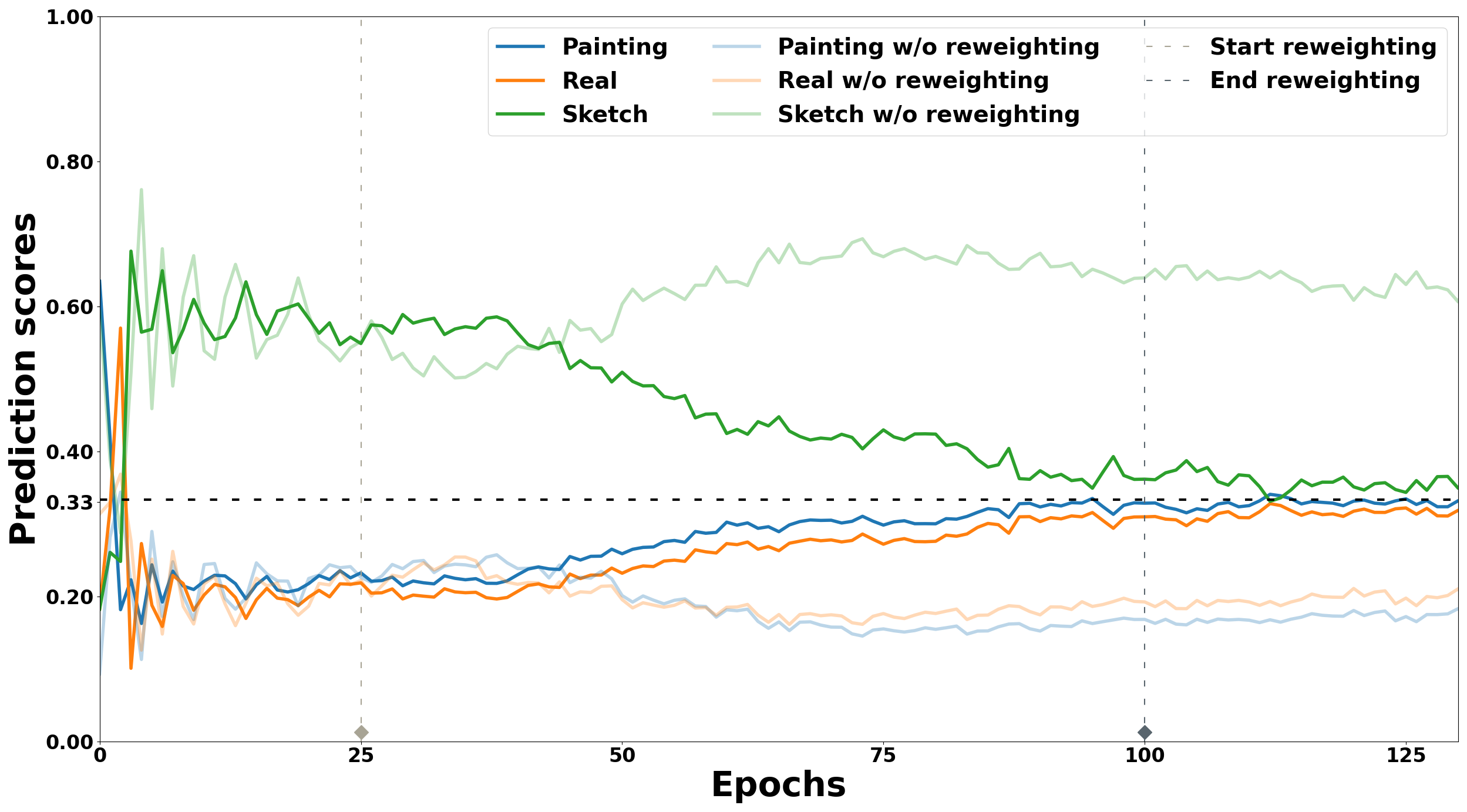}
    \vspace{-5pt}
    \caption{Study of invariance principle. 
    Illustration of prediction scores $p(x_i \in \mathcal{I}_{\text{Sketch}}|\mathbf{s}_i^0)$ estimated by domain classifier throughout training. The dashed line represents the Oracle score, illustrating a random guess \wrt domain category.
    }
    \label{fig:weights}
\end{figure}

To visualize the latent representation space and evaluate the effectiveness of two principles in DisMAE, we conduct a comprehensive set of experiments in DomainNet, employing Painting, Real, and Sketch domains as training domains and Clipart, Infograph, and Quickdraw for testing.
\begin{itemize}[leftmargin=*]
    \item \textbf{Reconstruction Visualization.} 
    Figures \ref{fig:intro} and more examples in Appendix \ref{sec:app_recon} demonstrate DisMAE's capability in image-level disentanglement.
    \item \textbf{Domain-Invariance Assessment.} 
    Figure \ref{fig:weights} emphasizes the domain-invariant proficiency of our semantic encoder, as evidenced by the prediction scores of $\mathbf{s}_x^0$.
    \item \textbf{Representation-level Visualization.}
    Through t-SNE visualizations in Figures \ref{fig:tsne_MAE} and \ref{fig:tsne_DisMAE}, we validate that DisMAE adheres to invariance and disentanglement principles at the representation level.
\end{itemize}
The specific analysis is detailed below:


\textbf{Disentanglement principle: Semantic and variational features learned by DisMAE are fully disentangled.}
From an image perspective, as illustrated in Figure \ref{fig:intro}, DisMAE differentiates between the foreground and background of an image, exemplified by its capability to transform the background to a blue sky or a red hue without affecting the primary subject, the flower. 
Remarkably, DisMAE can discern domain styles and fuse domain-specific elements across them — a notable instance is superimposing the sun from a sketch onto a painting. 
Such disentanglement ability endows DisMAE with the flexibility to generate controllable images by manipulating semantic and variation factors through swapping.
From a feature space viewpoint, as evidenced in Figure \ref{fig:tsne_DisMAE}, closely situated semantic representations display shared content features.
In contrast, in the space extracted by the variation encoder, neighboring data points share analogous attributes like color, texture, and background. 
\hb{It is worth noting that Figure \ref{fig:tsne_DisMAE}(b) shows some overlap between the painting and real domains. This phenomenon may arise due to the high similarity of images in these two domains.}

\textbf{Invariance Principle: Representations learned by semantic encoder are domain-invariant.}
As depicted in Figure \ref{fig:weights}, the initial prediction scores of Sketch for each domain (opaque lines), denoted as $p(x_i \in \mathcal{I}_{\text{Sketch}}|\mathbf{s}_i^0)$, exhibit randomness at the beginning. 
With the advancement of training, these prediction scores elevate for the Sketch domain (opaque green line), demonstrating that Sketch-specific features are extracted by the semantic encoder at first. As training progresses, the prediction scores gradually converge to 0.33 (domain-agnostic score), indicating domain-variant information excluded from our main encoder and the efficacy of our invariance principle. Conversely, in scenarios where the reweighting term is omitted, the prediction score for the Sketch domain (transparent green line) fails to converge to 0.33, revealing a domain-variant encoder. Overall, the figure further validates the impact of our adaptive reweighting term and the goal of the invariance principle. 
For more analysis, see Appendix \ref{sec:app_invariance}.


\subsection{Study on DisMAE (RQ3)}\label{sec:exp_rq3}

\begin{table}[t]
\centering
    \resizebox{0.55\linewidth}{!}{
   \begin{tabular}{c c|c c c|c c}
   \toprule
    \multicolumn{2}{c|}{} & \textbf{Clip} & \textbf{Info} & \textbf{Quick} & \textbf{Overall} & \textbf{Avg.} \\
    \midrule 
    \multicolumn{2}{c|}{MAE}&61.04&19.15&27.50&32.02&35.90 \\
    \multicolumn{2}{c|}{Inter-domain neg.}&61.82&17.60&31.60&33.91&37.01 \\
    \multicolumn{2}{c|}{w/o weights}&\underline{64.16}&17.60&\underline{34.40}&\underline{35.86}&\underline{38.72} \\
    \multicolumn{2}{c|}{Random weights}&\textbf{64.68}&\underline{19.15}&31.10&34.65&38.31 \\
    \multicolumn{2}{c|}{Reverse weights}&60.52&18.57&29.90&33.02&36.33 \\
    \midrule
    \multicolumn{2}{c|}{\textbf{DisMAE}}&63.12&\textbf{19.16}&\textbf{35.90}&\textbf{36.89}&\textbf{39.39} \\
     \bottomrule
    \end{tabular}}
    \vspace{5pt}
    \caption{Effect of adaptive contrastive loss $\Lapl_{\text{con}}$ on DisMAE.}    
\label{tab:ablation}
\vspace{-15pt}
\end{table}

To further evaluate each component of DisMAE, and its potential to enhance generalization ability, we conduct a comprehensive analysis using DomainBed in the UDG setting.
This analysis contains three-fold: an ablation study of the adaptive contrastive loss components, the impact of decoder depth, and the effect of mask ratios.
During the evaluation, we employ the Painting, Real, and Sketch domains as training domains and evaluate the models' performance on Clipart, Infograph, and Cartoon domains. 
Specifically, training is executed on ViT-Tiny/16 across 500 epochs, leveraging the full fraction of labeled data for fine-tuning. 
For thorough results and analysis, readers are referred to Appendix \ref{sec:app_ablation} due to space constraints.



\textbf{Effect of adaptive contrastive loss}. 
Table \ref{tab:ablation} shows that leveraging intra-domain negative samples significantly enhances generalization. 
The meticulously designed reweighting term aids in achieving the domain-invariance principle.
\section{Related work} \label{sec:app_related_work}
DisMAE is related to the literature on self-supervised learning (SSL), domain generalization (DG), and unsupervised domain generalization (UDG).

Self-supervised learning (SSL) employs a range of pretext tasks to derive meaningful representations from large amounts of unlabeled data, aiming to enhance performance in downstream tasks.
Contemporary SSL methods can roughly fall into two distinct research lines: contrastive approaches and generative models. 
Recent advancements in contrastive methods \cite{MoCo_V2, SimCLR, BYOL, SwAV, SimSiam, Barlow_twins, VICReg} have excelled in instance discrimination. 
They strategically draw two augmented versions of an image closer in the feature space, recognized as positives, while simultaneously distancing them from negatives.
While contrastive SSL methods have witnessed significant advancements, their efficacy often hinges on specific factors like high-quality data augmentation techniques and complicated strategies for training stabilization, such as well-designed negative sampling.
On the contrary, generative SSL can avoid these dependencies.
These approaches primarily focus on the reconstruction of intrinsic features and information.
In the field of computer vision (CV), inspired by the introduction of ViT \cite{ViT}, masked image modeling (MIM) has attracted the huge attention of the research community, such as bidirectional encoder representation from image transformer (BEiT) \cite{BEiT}, masked autoencoder (MAE) \cite{MAE}, context autoencoder (CAE) \cite{CAE}, a simple frame for MIM (SimMIM) \cite{SimMIM}, image generating pretraining (iGPT) \cite{iGPT}.
Diverging from the SSL methods previously discussed, which concentrate on single-domain learning, our study pivots towards a more novel and pragmatic task: pretraining across multiple domains.

Domain generalization (DG) aims to learn semantic representations that remain consistent across diverse domains, enabling models to generalize effectively to unseen domains using labeled data.
Common approaches for solving the DG task can be broadly categorized as follows: minimizing the difference among source domains by invariant learning \cite{IRM, VREx, IB-IRM}, adversarial learning \cite{CIAN, MADDG, ER}, or feature alignment \cite{Fishr, AdaRNN, DIFEX, Calibration}; modifying the inputs to assist in learning general representations \cite{CrossGrad, OpenDG, FSDR, SFA}; disentangling the features into domain-shared or domain-specific parts for better generalization \cite{DDG, mDSDI, DSR, DecAug, VDN}.

Unsupervised domain generalization (UDG) has been proposed recently as a more important and challenging task centered on pretraining with unlabeled source domains \cite{DARLING}. 
Research in UDG predominantly bifurcates into two methodologies: contrastive approaches and generative models.
In the realm of contrastive approaches, DARLING \cite{DARLING} pioneered the UDG framework, focusing on domain-aware representation learning with a novel contrastive loss. 
Another notable contribution is \cite{BrAD}, which innovatively conceptualized a bridge domain to unify all source domains. 
Additionally, DN$^2$A \cite{DN2A} advanced this line of work by implementing strong augmentations and reducing intra-domain connectivity.
Turning to generative models, DiMAE \cite{DiMAE} was instrumental in introducing cross-domain reconstruction tasks, where input images are augmented using style noise from varying domains. 
Building on this, CycleMAE \cite{cycleMAE} innovated with a cycle-structured cross-domain reconstruction task, in the absence of paired images. 
A novel attempt in this domain is our proposed DisMAE, which, to our knowledge, is the first to incorporate a disentangled generative pretraining approach within UDG.
\section{Conclusion}
Despite the great success of domain generalization tasks, the challenge of unsupervised domain generalization remains relatively underexplored.
In this work, we devised a principled disentangling approach, the disentangled mask autoencoder (DisMAE), to learn domain-invariant features in an unsupervised manner.
However, DisMAE does present a limitation meriting further exploration: its reconstruction accuracy is closely tied to the MAE backbone, which occasionally struggles to generate high-fidelity images (See failure examples in Appendix \ref{sec:app_failed}).

\section*{Acknowledgements}
This research is supported by the National Natural Science Foundation of China (92270114) and the advanced computing resources provided by the Supercomputing Center of the USTC.

%
%
\bibliographystyle{splncs04}
\bibliography{main}

\appendix
\clearpage

\section{Algorithm} \label{sec:app_algorithm}
Algorithm \ref{alg:DisMAE} depicts the detailed procedure of DisMAE.
\begin{algorithm}[h]
   \caption{DisMAE: Disentangling Masked Autoencoder for Unsupervised Domain Generalization}
   \label{alg:DisMAE}
\begin{algorithmic}
   \STATE {\bfseries Input:} $\mathcal{D}_{S}=\{(x_1,d_1),...,(x_n,d_n)\}$, maximum adaptive training epochs $E_{ad}$, maximum training epoch $E$, adaptive training intervals $T_{ad}$, batch size $B$, margin $\gamma$, coefficients $\lambda_1$, current training epoch $e$
   \STATE {\bfseries Initial:} Parameters of DisMAE (i.e parameter $\theta_s$, $\theta_v$, $\theta_{\mathcal{G}}$, and $\theta_{cls}$ for semantic encoder $\phi_s$, variation encoder $\phi_v$, decoder $g$, and domain classifier $f_{cls}$ ), $e \leftarrow 1$

   \REPEAT
   \STATE Freeze parameters of the domain classifier $\theta_{1}$ \\
   \STATE Compute $l(x_i,x_{ii}')$ and $p(x_i \in \mathcal{I}_{d_i}|\mathbf{s}_i^0)$ with $\theta_s, \theta_v, \theta_{\mathcal{G}}, \theta_1$ \\
   \STATE Compute $\Lapl_{\text{rec}} = \frac{1}{B}\sum_{i=1}^{B}\max \{||x_i - g(\mathbf{s}_i, \mathbf{v}_i^0)||_{l_2} - \gamma, \, 0\}$ \\
   \STATE Compute $\Lapl_{\text{con}}= \sum_{i=1}^{B} \frac{1}{p(x_i\in \mathcal{I}_{d_i}|\mathbf{s}_i^0)} \cdot l(x_i,x_{ii}')$ \\
   
    \STATE Compute $\Lapl=\Lapl_{\text{rec}}+\lambda_1\Lapl_{\text{con}}$ \\
    \STATE Update $\theta_s, \theta_v, \theta_{\mathcal{G}}$ by minimizing $\Lapl$

   \IF{ $e \bmod T_{ad} == 0$ and $ e \leq E_{ad}$}
    \STATE Freeze parameters of backbones $\theta_s$, $\theta_v$ and $\theta_{\mathcal{G}}$\\
    \STATE Compute $\Lapl_{cls}=\sum_{i=1}^{B} \frac{1}{B} \cdot CE(d_i, \hat{d}_i=f_{cls}(\mathbf{s}_i^0))$ \\
    \STATE Update $\theta_{cls}$ by minimizing $\Lapl_{cls}$ \\
   \ENDIF

   \STATE $e \leftarrow e+1$
   
   \UNTIL{$e == E$}
\end{algorithmic}
\end{algorithm}

\section{Discussion About Differences} \label{sec:app_novelty}

We argue that DisMAE is novel and significantly different from prior studies \wrt three aspects.
1) \textbf{Scope.}
Transitioning to UDG is non-trivial.
Previous disentangled methods like DADA~\cite{DADA}, DIDN~\cite{DIDN}, and DIR~\cite{DIR}, while effective in DG, struggle with unsupervised data due to their high dependence on class labels to encapsulate semantic attributes.
2) \textbf{Disentangled Targets.} 
Without class label guidance, achieving a domain-invariant semantic encoder is challenging.
Many UDG methods, such as DiMAE~\cite{DiMAE} and CycleMAE~\cite{cycleMAE}, can only separate domain styles using multiple decoders but fall short in disentangling domain-invariant semantics from variations.
3) \textbf{Disentangle Strategy}.
DisMAE is grounded in disentanglement and invariance principles, uniquely combining adaptive contrastive loss with reconstruction loss collaboratively.
The adaptive contrastive loss, in particular, is designed by seamlessly leveraging the domain classifier and intra-domain negative sampling.
The differences are summarized in Table \ref{tab:novelty}.

\begin{table}[h]
\centering   
\resizebox{0.99\linewidth}{!}{
    \begin{tabular}{c|ccccc}
    \toprule
    \textbf{} & \textbf{DADA}~\cite{DADA} & \textbf{DIDN}~\cite{DIDN}  & \textbf{DIR}~\cite{DIR} &\textbf{DiMAE}~\cite{DiMAE} & \textbf{DisMAE} \\
    \midrule
    \textbf{\makecell[c]{Class Label}}  & \Checkmark & \Checkmark & \Checkmark & \XSolidBrush & \XSolidBrush \\ \midrule
    \textbf{\makecell[c]{Disentanglement\\\&Invariance Strategy}} & \makecell[c]{Adversarial Training} & {Adversarial Training} & {Adversarial Training} & Fourier Transform & \makecell[c]{Adaptive Contrastive\\Learning}  \\ \midrule
    \textbf{Disentangled Info} & Semantics-Variations & - & Semantics-Variations & Domain Styles & Semantics-Variations \\ \midrule
    \textbf{Architecture} & \makecell[c]{1 Encoder\\1 Decoder} & \makecell[c]{2 Encoders\\1 Decoder} & \makecell[c]{2 Encoders\\1 Decoder\\1 GAN} & \makecell[c]{1 Encoder\\$N_d$ Decoders} & \makecell[c]{2 Encoders\\1 Decoder} \\
    \bottomrule
    \end{tabular}}
    \vspace{5pt}
    \caption{Comparison with previous works}
    \label{tab:novelty}
    
\end{table}

\section{Experiments} \label{sec:app_experiments}

\subsection{Experimental Settings} \label{sec:app_experiment_settings}


\textbf{Baseline Hyperparameter Tuning.}
For a fair comparison, we uniformly substitute the backbones of all baselines with the same ViT-B/16 and rerun the experiment using UDG and DG open-source codebases. And we provide the default hyperparameters for UDG baselines in Table \ref{tab:app_UDG_baselines}. And the search distribution for each hyperparameter in each DG baseline is detailed in Table \ref{tab:app_DG_baselines}.

\begin{table}[h]
\centering   
\resizebox{0.80\linewidth}{!}{
    \begin{tabular}{ll}
    \toprule
    \multicolumn{1}{c}{} & \multicolumn{1}{c}{Default hyperparameters}  \\
    \midrule
    MoCo V2 & lr=5e-4, BS=96, WD=0.05, K=65536, m=0.999, T=0.07 \\ 
    BYOL    & lr=5e-4, BS=64, WD=0.05, m=0.996 \\ 
    MAE     & lr=1e-3, BS=96, WD=0.00, mask ratio=0.75 \\ 
    DARLING & lr=5e-4, BS=96, WD=0.05, K=65536, m=0.995, T=0.07 \\ 
    CycleMAE& lr=7e-4, BS=96, WD=0.05, m=0.999, $\alpha=2$, $\beta=1$  \\
    \bottomrule
    \end{tabular}}
    \vspace{5pt}
    \caption{Hyperparameters for baselines in UDG. BS represents the batch size, and WD denotes weight decay.}
    \label{tab:app_UDG_baselines}
    
\end{table}

\begin{table*}[h]
\centering  
\resizebox{0.80\linewidth}{!}{
    \begin{tabular}{llll}
    \toprule
    \textbf{Condition}                       & \textbf{Parameter} & \textbf{Default value} & \textbf{Random distribution}  \\ 
    \midrule
    \multicolumn{1}{l}{\multirow{4}{*}{ViT}} & learning rate              & 0.00005 & $10^{Uniform(-5,-3.5)}$ \\ 
    \multicolumn{1}{l}{}                    & batch size                  & 32      & $2^{Uniform(3,5.5)}$    \\
    \multicolumn{1}{l}{}                    & generator learning rate     & 0.00005 & $10^{Uniform(-5,-3.5)}$  \\
    \multicolumn{1}{l}{}                    & discriminator learning rate & 0.00005 & $10^{Uniform(-5,-3.5)}$ \\
    \midrule
    MMD                                     & gamma                       & 1       & $10^{Uniform(-1,1)}$  \\
    \midrule
    GroupDRO                                & eta                          & 0.01   & $10^{Uniform(-3,-1)}$  \\
    \midrule
    \multicolumn{1}{l}{\multirow{2}{*}{IRM}} & lambda                        & 100   & $10^{Uniform(-1,5)}$  \\
    \multicolumn{1}{l}{}                    & iterations of penalty annealing & 500 &  $10^{Uniform(0,4)}$  \\
    \midrule
    Mixup & alpha & 0.2 & $10^{Uniform(0,4)}$  \\
    \midrule
    \multicolumn{1}{l}{\multirow{2}{*}{VREx}}& lambda                        & 10   & $10^{Uniform(-1,5)}$ \\
    \multicolumn{1}{l}{}                     & penalty anneal iters          & 500  & $10^{Uniform(0,4)}$ \\
    \midrule
    \multicolumn{1}{l}{\multirow{3}{*}{Fishr}} & lambda                     & 1000  & $10^{Uniform(1,4)}$ \\
    \multicolumn{1}{l}{}                    & penalty anneal iters          & 1500  & $Uniform(0,5000)$ \\
    \multicolumn{1}{l}{}                    & ema                           & 0.95  & $Uniform(0.90,0.99)$ \\
    \midrule
    \multicolumn{1}{l}{\multirow{12}{*}{DDG}}& recon w                    & 0.5  & $Random(0.1, 0.2, 0.5, 1.0)$ \\
    \multicolumn{1}{l}{}                     & recon x w                     & 0.5    & $Random(1, 2, 5, 10)$ \\
    \multicolumn{1}{l}{}                     & margin                        & 0.25  & $Random(0.10, 0.25, 0.5, 0.75)$ \\
    \multicolumn{1}{l}{}                     & recon xp w                     & 0.5    & $Random(1, 2, 5, 10)$ \\
    \multicolumn{1}{l}{}                     & recon xn w                        & 0.50  & $Random(0.10, 0.25, 0.5, 0.75)$ \\
    \multicolumn{1}{l}{}                     & max cyc w                     & 2.0    & $Random(1, 2, 4)$ \\
    \multicolumn{1}{l}{}                     & max w                        & 2.0  & $Random(0.5, 1.0, 2.0)$ \\
    \multicolumn{1}{l}{}                     & gan w                    & 1.0    & $Random(0.5, 1.0, 2.0)$ \\
    \multicolumn{1}{l}{}                     & eta                        & 0.01  & $Random(0.01, 0.05)$ \\
    \multicolumn{1}{l}{}                     & recon cyc w                    & 0.0    & $Random(0.1, 0.2, 0.5, 1.0)$ \\
    \multicolumn{1}{l}{}                     & warm iter r                     & 0.2  & $Random(0.1, 0.2, 0.3, 0.4, 0.5)$ \\
    \multicolumn{1}{l}{}                     & warm scale                     & 0.005    & $10^{Uniform(-5, -3)}$ \\
    
    \bottomrule
    \end{tabular}}
    \vspace{5pt}
    \caption{Default hyperparameters and random search distribution for baselines in DG}
    \label{tab:app_DG_baselines}
\end{table*}

\subsection{Overall performance} \label{sec:app_performance}

\textbf{Unsupervised Domain Generalization}. Due to limited space in the paper, we show the rest UDG results in Table \ref{tab:UDG2}. We employ Clipart, Infograph, and Quickdraw as training domains and Painting, Real, and Sketch as test domains. Following the same all correlated settings and protocols in DARLING, we find that our DisMAE could achieve 1.14\%, 1.19\%, 4.40\%, and 5.45\% gains for average accuracy over the second-best baselines across 1\%, 5\%, 10\%, and 100\% fraction setting respectively.

\begin{table*}[h]
\centering  
   \resizebox{0.90\linewidth}{!}{
    \begin{tabular}{c c c|c c c|c c|c c c|c c}
   \toprule
   & & & \multicolumn{5}{c|}{\textbf{Label Fraction 1\% (Linear evaluation)}} & \multicolumn{5}{c}{\textbf{Label Fraction 5\% (Linear evaluation)}} \\
    \midrule
    \multicolumn{3}{c|}{\textbf{Methods}} & \textbf{Paint} & \textbf{Real} & \textbf{Sketch} & \textbf{Overall} & \textbf{Avg.} & \textbf{Paint} & \textbf{Real} & \textbf{Sketch}  & \textbf{Overall} & \textbf{Avg.} \\
    \midrule
    \multicolumn{3}{c|}{ERM \cite{ERM}}&7.98&9.94&5.38&8.46&7.77&6.48&8.64&9.84&8.29&8.32 \\
    \multicolumn{3}{c|}{MoCo V2 \cite{MoCo_V2}} &7.34&10.19&4.46&8.22&7.33&11.15&13.28&7.09&11.41&10.51 \\
    \multicolumn{3}{c|}{BYOL \cite{BYOL}}&\underline{9.23}&\underline{10.57}&10.58&10.21&10.13&\underline{12.43}&17.34&14.58&15.44&14.78 \\
    \multicolumn{3}{c|}{MAE \cite{MAE}} &8.80&10.30&\textbf{12.62}&\underline{10.38}&\underline{10.57}&12.13&\underline{17.63}&\underline{15.02}&\underline{15.60}&\underline{14.93} \\
    \multicolumn{3}{c|}{DARLING \cite{DARLING}}&8.59&9.01&11.10&9.32&9.57&9.31&12.00&13.72&11.61&11.68 \\
    \multicolumn{3}{c|}{CycleMAE \cite{cycleMAE}} &9.21&9.49&6.62&8.82&8.44&11.44&14.21&10.01&12.58&11.88 \\
    \midrule
    \multicolumn{3}{c|}{\textbf{DisMAE (Ours)}}&\textbf{11.04}&\textbf{11.64}&\underline{12.46}& \textbf{11.65}&\textbf{11.71}&\textbf{13.69}&\textbf{17.74}&\textbf{16.92}&\textbf{16.47}&\textbf{16.12} \\
    \midrule
    & & & \multicolumn{5}{c|}{\textbf{Label Fraction 10\% (Full finetuning)}} & \multicolumn{5}{c}{\textbf{Label Fraction 100\% (Full finetuning)}} \\
    \midrule
    \multicolumn{3}{c|}{\textbf{Methods}} & \textbf{Paint} & \textbf{Real} & \textbf{Sketch} & \textbf{Overall} & \textbf{Avg.} & \textbf{Paint} & \textbf{Real} & \textbf{Sketch}  & \textbf{Overall} & \textbf{Avg.} \\
    \midrule
    \multicolumn{3}{c|}{ERM \cite{ERM}} &14.14&16.76&12.63&15.19&14.51&26.25&33.29&23.25&29.28&27.59 \\
    \multicolumn{3}{c|}{MoCo V2 \cite{MoCo_V2}} &16.45&19.12&9.48&16.39&15.02&22.94&35.26&18.14&28.36&25.44 \\
    \multicolumn{3}{c|}{BYOL \cite{BYOL}} &19.12&20.55&18.68&19.77&19.45&24.91&35.72&24.19&30.39&28.27 \\
    \multicolumn{3}{c|}{MAE \cite{MAE}} &\underline{19.34}&\underline{23.07}&\underline{24.18}&\underline{22.29}&\underline{22.20}&\underline{31.72}&\underline{43.72}&\underline{36.74}&\underline{39.02}&\underline{37.40} \\
    \multicolumn{3}{c|}{DARLING \cite{DARLING}} &13.72&19.76&16.43&17.40&16.64&25.87&37.60&26.67&32.11&30.05 \\
    \multicolumn{3}{c|}{CycleMAE \cite{cycleMAE}} &17.80&22.93&17.15&20.23&19.29&31.72&39.44&30.00&35.39&33.72 \\
    \midrule
    \multicolumn{3}{c|}{\textbf{DisMAE (Ours)}} &\textbf{24.49}&\textbf{27.06}&\textbf{28.24}&\textbf{26.61}&\textbf{26.60}& \textbf{38.89}&\textbf{45.95}&\textbf{43.72}&\textbf{43.58}&\textbf{42.85} \\
    \bottomrule
    \end{tabular}}
    \vspace{5pt}
    \caption{Unsupervised domain generalization results on DomainNet. We employ Clipart, Infograph, and Quickdraw as training domains and Painting, Real, and Sketch as test domains. All the models are unsupervised pre-trained before fine-tuning on the labeled data. Overall and Avg. are the overall test data accuracy and the arithmetic mean of individual domain accuracy respectively. Note that they are different because the size of each test domain isn’t equal. \textbf{Bold}=best, \underline{underline}=second best.}
    \label{tab:UDG2}
\end{table*}

\textbf{Domain Generalization}. Aligning with the training-domain validation setup in DomainBed, we achieve 0.50\% gains for the average accuracy in PACS datasets, as shown in Table \ref{tab:PACS}.

\begin{table*}[t]
\centering   
\resizebox{0.50\linewidth}{!}{
   \begin{tabular}{c c c|c c c c|c}
   \toprule
    \multicolumn{3}{c|}{\textbf{Methods}}& \textbf{A} & \textbf{C} & \textbf{P} & \textbf{S} & \textbf{Avg.} \\
    \midrule 
    \multicolumn{3}{c|}{ERM \cite{ERM}}&85.34&80.26&96.83&71.75&83.55 \\
    \multicolumn{3}{c|}{MMD \cite{MMD}}&84.83&79.53&96.16&75.42&83.98 \\
    \multicolumn{3}{c|}{GroupDRO \cite{GroupDRO}}&83.99&79.96&97.36&76.38&84.42 \\
     \multicolumn{3}{c|}{IRM \cite{IRM}}&\underline{86.05}&81.86&97.33&70.00&83.81 \\
     \multicolumn{3}{c|}{Mixup \cite{Mixup}}&85.27&80.70&97.13&72.77&83.97 \\
     \multicolumn{3}{c|}{VREx \cite{VREx}}&83.51&82.71&\textbf{97.88}&\underline{76.55}&85.16 \\
     \multicolumn{3}{c|}{Fishr \cite{Fishr}}&84.60&80.06&\underline{97.68}&72.40&83.69 \\
     \multicolumn{3}{c|}{DDG \cite{DDG}}&84.23&82.29&97.63&76.41&84.69 \\
     \multicolumn{3}{c|}{RIDG \cite{RIDG}}&\textbf{86.09}&\textbf{84.81}&96.41&76.18&\underline{85.87} \\
     \midrule
     \multicolumn{3}{c|}{\textbf{DisMAE (Ours)}}&85.48&\underline{83.19}&97.16&\textbf{79.66}&\textbf{86.37} \\
     \bottomrule
    \end{tabular}}
    \vspace{5pt}
\caption{Domain generalization results on PACS. \textbf{Bold}=best, \underline{underline}=second best.}
\label{tab:PACS}
\end{table*}

\subsection{Discussion of the invariance principle} \label{sec:app_invariance}
In Figure \ref{fig:tsne_MAE} and \ref{fig:tsne_DisMAE}, we visualize the representation acquired through MAE, our semantics encoder, and the variational encoder via t-SNE. We find that: (1) The representations generated by MAE for each domain showcase a degree of overlap at the center of the picture, accompanied by slight variations within each distribution. This suggests that MAE captures both semantics and domain-variant information but fails to disentangle them effectively. (2) Our semantic representations in each domain distribute uniformly. This justifies that DisMAE could learn domain-invariant representations from each domain. (3) The variation representations in each domain has their specific distribution. Clusters of similar variation data further emphasize domain-specific characteristics.

\subsection{More ablation study} \label{sec:app_ablation}

\textbf{Effects of decoder depth}. The efficacy of the adaptive contrastive loss hinges on the output of decoders. This prompts the inquiry: how many decoder layers are optimal for achieving peak performance? As shown in Table \ref{tab:depth}, a deeper decoder may lead to overfitting in reconstruction and subsequently diminish the effect of our contrastive loss. Thus, adopting a lightweight decoder could both accelerate the training and guarantee robustness.

\begin{table}[h]
\centering   
   \begin{tabular}{c c | c c}
    \multicolumn{2}{c|}{\textbf{Decoder Layer}} & \textbf{Overall} & \textbf{Avg.} \\
    \hline
    \rowcolor{lightgray} \multicolumn{2}{c|}{1}&\textbf{36.89}&\textbf{39.39} \\
    \multicolumn{2}{c|}{2}&34.70&37.99 \\
    \multicolumn{2}{c|}{4}&34.49&37.08 \\
    \multicolumn{2}{c|}{8}&32.97&36.59 \\
    \end{tabular}
    \vspace{5pt}
    \caption{Hyperparameter Analysis of the decoder layer}
\label{tab:depth}
\end{table}

\textbf{Effects of mask ratios}. In Table \ref{tab:mask_ratio}, we set different mask ratios to test the robustness of our model. And we found that the 80 percentile of the mask ratio reaches the optimal result. We set it as our default protocol.

\begin{table}[h]
\centering   
   \begin{tabular}{c c | c c}
    \multicolumn{2}{c|}{\textbf{Mask Ratio}} & \textbf{Overall} & \textbf{Avg.} \\
    \hline
    \multicolumn{2}{c|}{50}&33.07&36.61 \\
    \multicolumn{2}{c|}{60}&33.23&36.29 \\
    \multicolumn{2}{c|}{70}&34.20&37.80 \\
    \rowcolor{lightgray} \multicolumn{2}{c|}{80}&\textbf{36.89}&\textbf{39.39} \\
    \multicolumn{2}{c|}{90} & 34.91 & 37.20 \\
    \end{tabular}
    \vspace{5pt}
    \caption{Hyperparameter Analysis of the mask ratio}
\label{tab:mask_ratio}
\end{table}

\subsection{Failure cases} \label{sec:app_failed}

\begin{figure*}[t]
    \centering
    \includegraphics[width=0.55\linewidth]{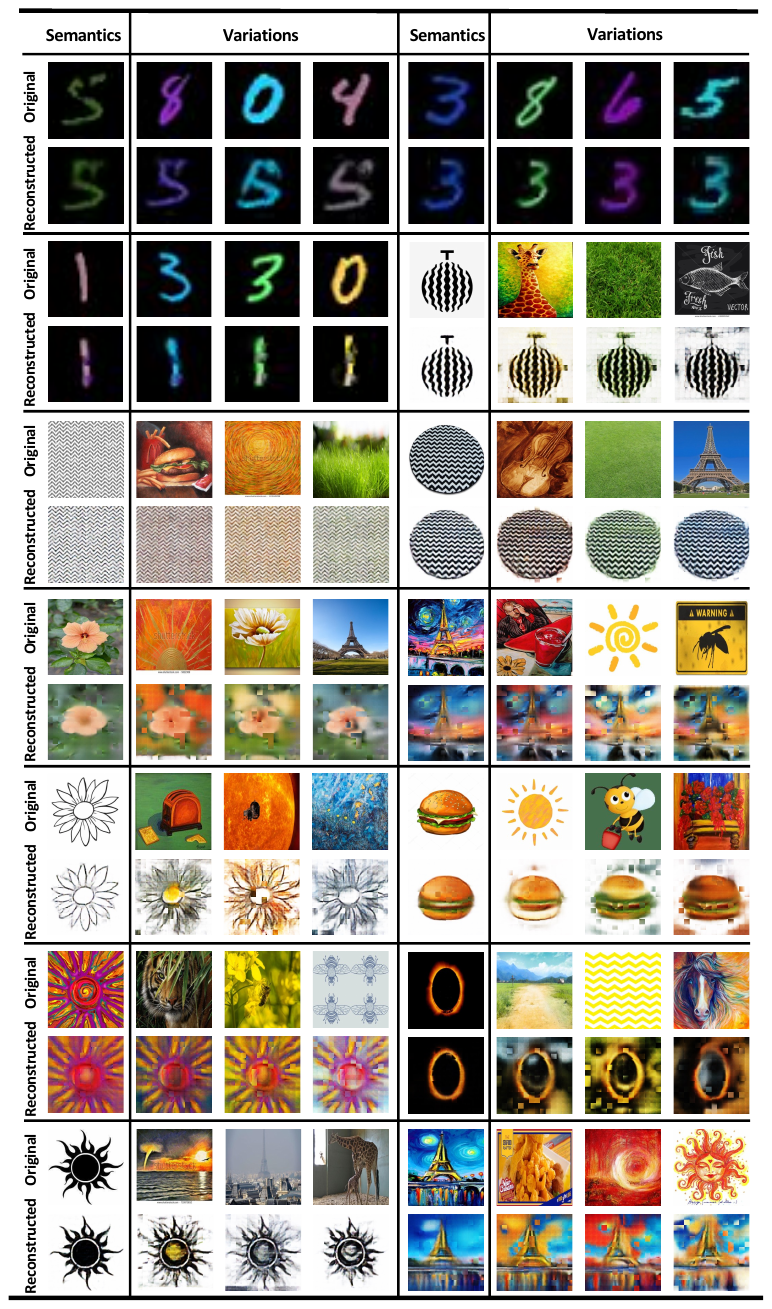}
    \caption{Some failure cases of reconstructed images generated by DisMAE.
    }
    \label{fig:failed}
\end{figure*}

Some failure cases of our proposed DisMAE are in Figure \ref{fig:failed}.
Our approach struggles with reconstruction containing intricate details and lines. 
It frequently fails to generate images that possess sufficient detail while simultaneously providing clear augmented variations. 
We attribute these failures to two primary reasons: 1) The MAE backbone operates on patches, making pixel-level reconstruction difficult, and our method heavily relies on the MAE model's reconstruction outcomes. 2) Our disentanglement lacks granularity, often capturing broad color regions and background information rather than nuanced details.
In the context of UDG, reconstructing images with fine granularity, high resolution, and authenticity remains a challenging and crucial research direction. 
We are also keenly interested in exploring the potential integration of the diffusion model within the UDG framework.

\subsection{Qualitative reconstructions} \label{sec:app_recon}

\begin{figure*}[h]
    \centering
    \includegraphics[width=0.8\linewidth]{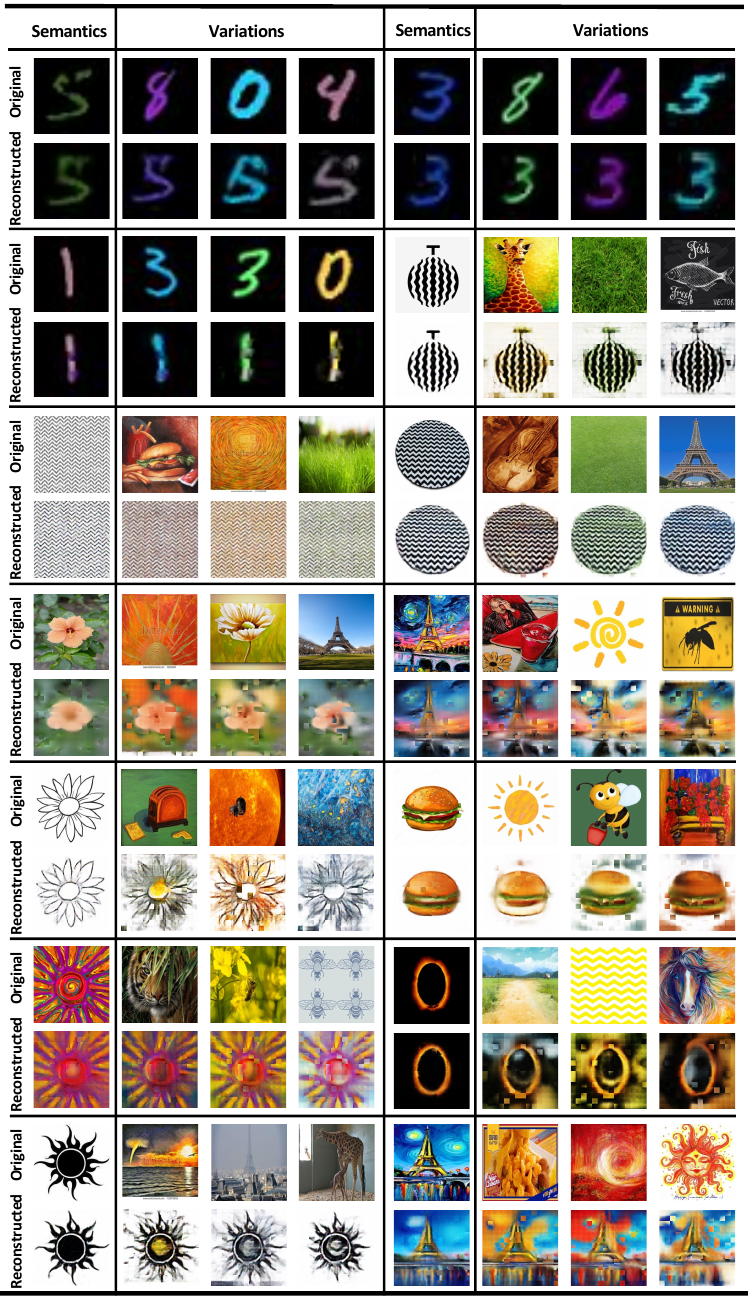}
    \caption{Illustrative reconstructed images generated by DisMAE.
    }
    \label{fig:more_recon}
\end{figure*}

Additional visualization results of image reconstruction, spanning both colored MNIST and DomainNet, can be observed in Figure \ref{fig:more_recon}.

DisMAE differentiates between the foreground and background of an image. 
Remarkably, DisMAE can discern domain styles and fuse domain-specific elements across them — a notable instance is superimposing the sun from a sketch onto a painting. 
Such disentanglement ability endows DisMAE with the flexibility to generate controllable images by manipulating semantic and variation factors through swapping.

\subsection{Detailed implementation of DisMAE} \label{sec:app_implementation}
We conduct all the experiments in Pytorch on a cluster of 8 NVIDIA Tesla A100 GPUs with 40GB each. Our default backbone consists of 12 blocks of semantic encoder, 6 blocks of variation encoder, and a transform-based decoder. We utilize ViT-B/16 as our default backbone for both visualization and main experiments. And we use ViT-Tiny/16 in our ablation study. We let margin $\gamma=0.008$ and $\tau=0.4$. In UDG, we choose the AdamW optimizer for the main branch and set the learning rate as 1e-4 and betas as (0.9, 0.95) for pre-training. As for finetuning, we adopt the learning rate as 0.025, 0.05, 5e-5, 5e-5 and batch size as 96, 192, 36, and 36 in the label fraction 1\%, 5\%, 10\%, and 100\% experiments respectively. And we finetune all the checkpoints for 50 epochs. In DG, $\lambda_1$ is selected within \{5e-4, 1e-3, 5e-3, 1e-2\} and $\lambda_2$ is selected within \{0.1, 0.5, 1.0, 2.0\}. The detailed hyperparameters for UDG and DG are listed in Table \ref{tab:app_implementation}.

\textbf{Details about training the domain classifier}. As for the domain classifier, we use the SGD optimizer with a learning rate of 0.0005, momentum of 0.99, and weight decay of 0.05. We choose adaptive training intervals $T_{ad}$ as 15 and maximum adaptive training epochs $E_{ad}$ as 100 in the UDG setting. We only update the domain classifier by minimizing the cross-entropy loss while freezing backbones when $e \bmod T_{ad} == 0$ and $ e \leq E_{ad}$, where $e$ is the current training epoch. The detailed algorithm can be found in Appendix~\ref{sec:app_algorithm}.

\begin{table*}[t]
   \centering
   \resizebox{0.70\linewidth}{!}{
   \begin{tabular}{c | c c c c c}
   \toprule
                                                        & batch size & weight decay & $\lambda_1$ & $\lambda_2$ & mask ratio \\ \midrule
    \textbf{UDG}                                        &         \multicolumn{5}{c}{\textbf{DomainBed}} \\ \midrule
    Paint, Real, Sketch $\rightarrow$ Clip, Info, Quick & $ 96 \times 3$ & 0.05        &  1e-3      &  0          & 0.80       \\ 
    Clip, Info, Quick $\rightarrow$ Paint, Real, Sketch & $ 96 \times 3$ & 0.05        &  1e-3      &  0          & 0.80      \\ \midrule
    \textbf{DG}                                         & \multicolumn{5}{c}{\textbf{PACS}} \\ \midrule
    C, P, S $\rightarrow$ A                               & $ 32 \times 3$ & 0.05        &  1e-3      &  1.0        & 0.55       \\
    A, P, S $\rightarrow$ C                               & $ 32 \times 3$ & 0.05        &  1e-3      & 1.0         & 0.40        \\  
    A, C, S $\rightarrow$ P                               & $ 32 \times 3$ & 0.05        &  5e-3      & 1.0         & 0.30        \\
    A, C, P $\rightarrow$ S                               & $ 40 \times 3$  & 0.05       &  1e-2      & 0.5         & 0.55         \\ \midrule 
      \textbf{DG}                                       & \multicolumn{5}{c}{\textbf{VLCS}} \\ \midrule
    L, S, V $\rightarrow$ C                               & $ 32 \times 3$ & 0.05        &  1e-3      &  1.0        & 0.70       \\
    C, S, V $\rightarrow$ L                               & $ 32 \times 3$ & 0.05        &  1e-3      & 1.0         & 0.40        \\  
    C, L, V $\rightarrow$ S                               & $ 32 \times 3$ & 0.05        &  5e-3      & 1.0         & 0.40        \\
    C, L, S $\rightarrow$ V                               & $ 40 \times 3$  & 0.05       &  1e-3      & 1.0         & 0.40         \\ \bottomrule
    \end{tabular}}
    \vspace{5pt}
    \caption{Hyperparameters Selection of DisMAE.}
\label{tab:app_implementation}
\end{table*}

\end{document}